\documentclass[letterpaper]{article} 
\usepackage{aaai25}  
\usepackage{times}  
\usepackage{helvet}  
\usepackage{courier}  
\usepackage[hyphens]{url}  
\usepackage{graphicx} 
\usepackage{natbib}  
\usepackage{caption} 
\usepackage{algorithm}

\usepackage{newfloat}
\usepackage{listings}
\DeclareCaptionStyle{ruled}{labelfont=normalfont,labelsep=colon,strut=off} 
\floatstyle{ruled}
\newfloat{listing}{tb}{lst}{}
\floatname{listing}{Listing}
\usepackage{algpseudocode}
\usepackage{enumitem}
\usepackage{booktabs}
\usepackage{multirow}
\usepackage{amsmath}
\usepackage{amssymb}
\usepackage{bbding}
\usepackage{subfigure}
\usepackage{colortbl}
\usepackage{pifont}
\usepackage{makecell}
\usepackage{mathrsfs}
\usepackage{bm}
\usepackage{dsfont}
\usepackage{xcolor}
\usepackage{soul}
\usepackage{framed}
\definecolor{shadecolor}{rgb}{0.92,0.92,0.92}
\usepackage{tcolorbox}
\usepackage{tabularx}
\usepackage{tabulary}
\usepackage{footmisc}

\newcommand{\ourmethod}{\textsc{TableLLM}}
\newcommand{\benchmark}{TableBench}
\newcommand{\trainset}{TableInstruct}

\newcommand{\simtask}{TableQA}

\title{\benchmark{}: A Comprehensive and Complex Benchmark \\
for Table Question Answering}
\author {
    Xianjie Wu\textsuperscript{\rm 1},
    Jian Yang\textsuperscript{\rm 1 \thanks{Corresponding author.}},
    Linzheng Chai\textsuperscript{\rm 1},
    Ge Zhang\textsuperscript{\rm 2},
    Jiaheng Liu\textsuperscript{\rm 1},
    Xeron Du\textsuperscript{\rm 2},
    Di Liang\textsuperscript{\rm 3}, \\
    Daixin Shu\textsuperscript{\rm 1},
    Xianfu Cheng\textsuperscript{\rm 1},
    Tianzhen Sun\textsuperscript{\rm 1},
    Tongliang Li\textsuperscript{\rm 4},
    Zhoujun Li\textsuperscript{\rm 1 $^*$},
    Guanglin Niu\textsuperscript{\rm 1}
}
\affiliations {
    \textsuperscript{\rm 1}Beihang University \\
    \textsuperscript{\rm 2}M-A-P \\
    \textsuperscript{\rm 3}Fudan University \\
    \textsuperscript{\rm 4}Beijing Information Science and Technology University \\
    \{wuxianjie, jiaya, lizj\}@buaa.edu.cn
}

\begin{document}

\maketitle

\begin{abstract}
Recent advancements in large language models (LLMs) have markedly enhanced the interpretation and processing of tabular data, introducing previously unimaginable capabilities.
Despite these achievements, LLMs still encounter significant challenges when applied in industrial scenarios, particularly due to the increased complexity of reasoning required with real-world tabular data, underscoring a notable disparity between academic benchmarks and practical applications. 
To address this discrepancy, we conduct a detailed investigation into the application of tabular data in industrial scenarios and propose a comprehensive and complex benchmark \benchmark{}, including 18 fields within four major categories of table question answering (\simtask{}) capabilities.
Furthermore, we introduce \ourmethod{}, trained on our meticulously constructed training set \trainset{}, achieving comparable performance with GPT-3.5.
Massive experiments conducted on \benchmark{} indicate that both open-source and proprietary LLMs still have significant room for improvement to meet real-world demands, where the most advanced model, GPT-4, achieves only a modest score compared to humans.
\end{abstract} 

\begin{links}
\link{Code}{https://github.com/TableBench/TableBench}
\end{links}

\section{Introduction}
\label{sec:introduction}

Recent studies have shown the potential of large language models (LLMs) on tabular tasks such as table question answering (TableQA)~\citep{zhu2021tat,zhao2023large,hegselmann2023tabllm,li2023table,zhang2024tablellm,lu2024large} by adopting in-context learning and structure-aware prompts~\citep{singha2023tabular}, suggesting that a well-organized representation of tables improves the interpretation of tabular. ~\citet{tai2023exploring} notes that eliciting a step-by-step reasoning process from LLMs enhances their ability to comprehend and respond to tabular data queries. Furthermore, ~\citet{zha2023tablegpt} investigates the use of external interfaces for improved understanding of tabular data.

\begin{figure}[tbp]
\centering
\includegraphics[]{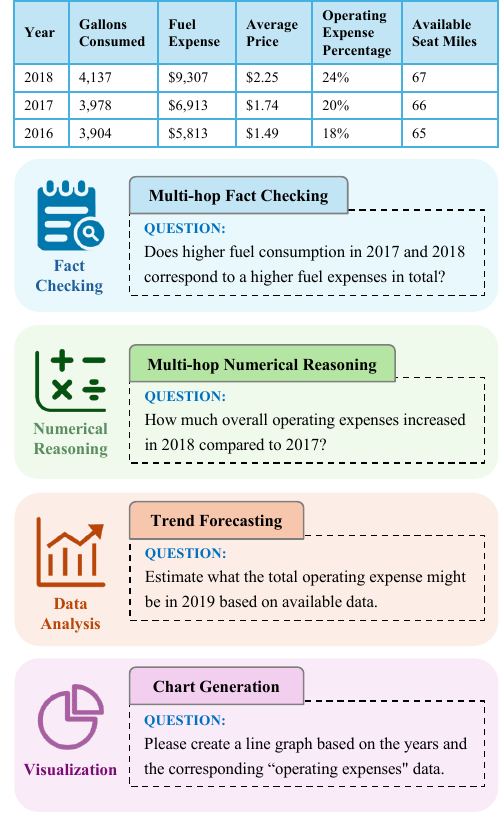}
\caption{
Typical challenges from four major categories (fact checking, numerical reasoning, data analysis, visualization) in \benchmark{}.}
\label{fig:intro-case}
\end{figure}

\begin{figure*}[htb]
\centering
\includegraphics[]{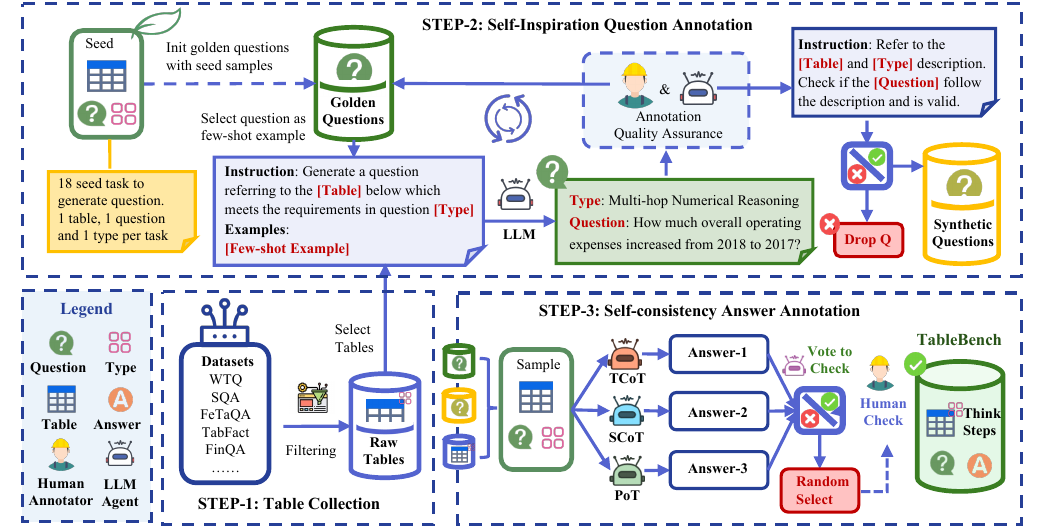}
\caption{A comprehensive overview of the annotation framework.}
\label{fig:dataset_construction}
\end{figure*}

Traditionally, adapting language models for tabular data processing entailed modifying their architectures with specialized features such as position embeddings and attention mechanisms to grasp structural nuances of tables. However, the introduction of LLMs like GPT-4, GPT-3.5~\citep{gpt3,gpt4}, and PaLM2~\citep{anil2023palm} has heralded a new approach focused on the art of crafting precise, information-rich prompts that seamlessly integrate table data, coupled with leveraging external programming languages like SQL, Python, or other languages~\citep{wang2023mac,chai2024mceval}, which facilitates more sophisticated chain-of-thought~\citep{cot} (CoT) reasoning processes across both proprietary and open-source LLM platforms, including Llama. Such advancements have propelled the fine-tuning of models for tabular data-specific tasks, showcased by initiatives like StructLM~\citep{zhuang2024structlm}, enhancing capabilities in table structure recognition, fact verification, column type annotation, and beyond. 
However, the existing benchmark might not entirely resonate with the practical challenges, especially complex reasoning requirements encountered by professionals routinely navigating tabular data in real-world settings. \textit{Therefore, there is a huge need for creating a benchmark to bridge the gap between the industrial scenarios and the academic benchmark.}

To better evaluate the capability of LLMs in Table QA, we introduce \benchmark{}, a comprehensive and complex benchmark covering 18 subcategories within four major categories of TableQA abilities, as illustrated in Figure \ref{fig:intro-case}. 
First, We systematically analyze real-world challenges related to table applications and define task complexity based on the required number of reasoning steps.
Based on the analysis, we introduce a rigorous annotation workflow, integrating manual and automated methods, to construct \benchmark{}.
Subsequently, We create a massively TableQA instruction corpora \trainset{}, covering three distinct reasoning methods.
Textual chain-of-thought (TCoT) utilizes a textual reasoning approach, employing a series of inferential steps to deduce the final answer.
Symbolic chain-of-thought (SCoT) adopts symbolic reasoning steps, leveraging programming commands to iteratively simulate and refine results through a \textit{Think then Code} process. 
Conversely, program-of-thought (PoT) generates executable code, using lines of code as reasoning steps within a programming environment to derive the final result. 
Based on open-source models and \trainset{}, we propose \ourmethod{} as a strong baseline to explore the reasoning abilities of LLMs among tabular data, yielding comparable performance with GPT-3.5.
Furthermore, we evaluate the performance of over 30 LLMs across these reasoning methods on \benchmark{}, highlighting that both open-source and proprietary LLMs require substantial improvements to meet real-world demands. Notably, even the most advanced model, GPT-4, achieves only a modest score when compared to human performance.

The contributions are summarized as follows: 
\begin{itemize}
    \item We propose \benchmark{}, a human-annotated comprehensive and complex TableQA benchmark comprising 886 samples across 18 fields, designed to facilitate fact-checking, numerical reasoning, data analysis, and visualization tasks. 
    \item  We introduce \trainset{}, a massive TableQA instruction corpus covering three distinct reasoning methods. \ourmethod{}, trained on \trainset{}, serves as a robust baseline for \benchmark{}.
    \item We systematically evaluate the interpretation and processing capabilities of more than 30 models on our crafted \benchmark{} and create a leaderboard to evaluate them on four main tasks. Notably, extensive experiments suggest that comprehensive and complex TableQA evaluation can realistically measure the gap between leading language models and human capabilities in real-world scenarios.
\end{itemize}

\section{Construction of \benchmark{}}
\label{sec:construction_of_tablebench}
To bridge the gap between academic benchmarks and industrial scenarios, we comprehensively analyze tabular data applications in real-world contexts, categorizing these problems into four major categories and 18 specific subcategories. We define the complexity of these tasks based on the reasoning steps required for problem-solving and provide detailed guidelines for defining and decomposing these steps, which are rigorously followed during the annotation process. Additionally, we introduce an annotation framework that combines manual and automated methods to enhance annotation efficiency, as illustrated in Figure \ref{fig:dataset_construction}. Finally, we propose two high-quality corpora: \benchmark{}, a comprehensive and complex benchmark consisting of 886 samples, and \trainset{} (20K samples in total), massive instruction corpora designed to instruct LLMs with various reasoning methods.

\begin{table}[tb]
\centering
\footnotesize \selectfont
\begin{tabular}{lr}
    \toprule
    \textbf{Properties} & \textbf{Value}\\
    \midrule
    \multicolumn{2}{c}{\textbf{\textit{Basic Insight}}} \\
     \midrule
    \textbf{Unique Tables} & 3681 \\
    \textbf{Question Length(Avg)} & 20.30 \\
    \textbf{Answer Length (Avg)} & 8.52 \\
    \textbf{Columns Per Table} & 6.68 \\
    \textbf{Rows Per Table } & 16.71 \\
    \textbf{Ratio of Numerical Cells} & 65.74\% \\
    \textbf{Average Reasoning Steps} & 6.26 \\
    \midrule
    \multicolumn{2}{c}{\textbf{\textit{Question Categories}}} \\
    \midrule
    \textbf{Fact Checking} & Match-Based Fact Checking  \\
                 &   Multi-hop Fact Checking  \\
    \textbf{Numerical Reasoning} & Arithmetic Calculation  \\
                &    Comparison  \\
                &    Aggregation  \\
                &    Ranking  \\
                &    Counting  \\
                &    Time-based Calculation  \\
                &    Multi-hop Numerical Reasoning  \\
                &    Domain-Specific  \\
    \textbf{Data Analysis} & Descriptive Analysis  \\
                & Anomaly Detection  \\
                & Statistical Analysis  \\
                & Correlation Analysis  \\
                & Causal Analysis  \\
                & Trend Forecasting  \\
                & Impact Analysis  \\
    \textbf{Visualization} & Chart Generation  \\
    \midrule
    \textbf{\benchmark{} Size} & 886 \\
    \textbf{\trainset{} Size} & 19,661 \\
    \bottomrule
\end{tabular}
\caption{Data statistics of \benchmark{}}
\label{table:dataset_statistic}
\end{table}

\subsection{Tabular Data Collection} 
\label{secsec:tabular_data_collection}
We collect raw tabular data from existing datasets, including typical datasets such as WTQ~\citep{pasupat2015compositional}, SQA~\citep{iyyer2017search}, TabFact~\citep{nan2022fetaqa}, FeTaQA~\citep{nan2022fetaqa}, FinQA~\citep{chen2021finqa}, AIT-QA~\citep{katsis2022ait}, etc. 
To align closely with the "\textit{reasoning complexity of questions}" dimension in real-world tabular problems, we do not specifically design for the complexity of the tables themselves, such as structural complexity or large-sized tables. Instead, we adopt a moderate complexity in tabular data.
We select tables based on topics and size, ensuring each contains at least 8 rows and 5 columns. We focus on tables with significant numerical values to emphasize numerical reasoning, thereby ensuring depth in numerical computation reasoning.
Ultimately, we collect 3681 tables covering 20 major topics: finance, competition, sports, science, etc.

\begin{figure}[tp]
\centering
\includegraphics[]{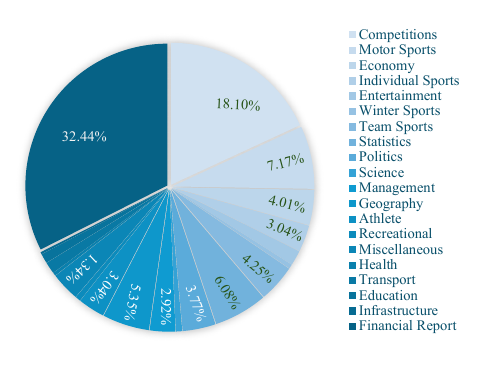}    
\caption{Topic distribution across tables}
\label{fig:table_topic_distribution}
\end{figure}

\begin{table}[bp]
\centering
\scriptsize \selectfont
\begin{tabular}{l|cccc}
\toprule
\textbf{\makecell[c]{Dataset}} & \textbf{\makecell[c]{Fact \\Checking}} & \textbf{\makecell[c]{Numerical \\Reasoning}} & \textbf{\makecell[c]{Data \\Analysis}} & \textbf{\makecell[c]{Visulization}}\\
\midrule
\textbf{WTQ} & \Checkmark & \ding{55} & \ding{55} & \ding{55}  \\
\textbf{SQA} & \Checkmark & \ding{55} & \ding{55} & \ding{55}  \\
\textbf{TabFact} & \Checkmark & \ding{55} & \ding{55} & \ding{55} \\
\textbf{FeTaQA} & \Checkmark & \ding{55} & \ding{55} & \ding{55} \\
\midrule
\textbf{FinQA} & \ding{55} & \Checkmark & \ding{55} & \ding{55} \\
\textbf{AIT-QA} & \ding{55} & \Checkmark & \ding{55} & \ding{55} \\
\textbf{WikiSQL} & \Checkmark & \Checkmark & \ding{55} & \ding{55} \\
\textbf{Spider} & \Checkmark & \Checkmark & \ding{55} & \ding{55} \\
\textbf{Bird} & \ding{55} & \Checkmark & \Checkmark & \ding{55} \\
\textbf{Text2Analysis} & \ding{55} & \ding{55} & \Checkmark & \Checkmark \\
\midrule
\textbf{\benchmark{}} & \Checkmark & \Checkmark & \Checkmark & \Checkmark \\
\bottomrule
\end{tabular}
\caption{Comparison with existing datasets in categories.}
\label{table:comparison_with_other_dataset}
\end{table}

\begin{figure*}[tp]
    \centering
    \subfigure[Reasoning steps of various question categories in TableBench]{
        \includegraphics[]{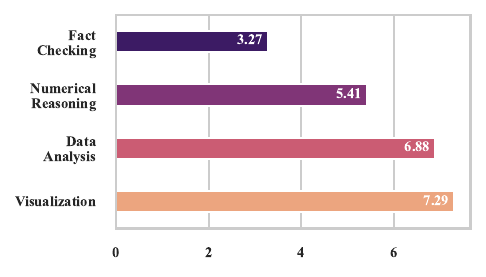} 
        \label{fig:subB}
    }
    \hfill 
    \subfigure[Reasoning steps comparison across different datasets]{
        \includegraphics[]{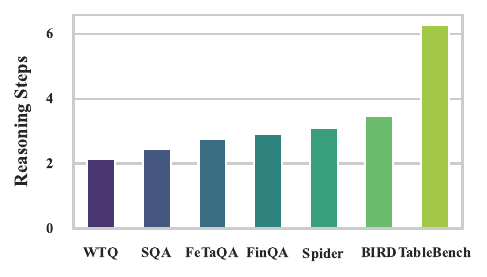} 
        \label{fig:subA}
    }
    \caption{Comparison with existing datasets in reasoning complexity.} 
    \label{fig:reasoning_steps} 
\end{figure*}

\subsection{Question Annotation}
\label{secsec:question_annotation}
We opt to manually construct a more complex set of questions to mitigate the data leak risk in LLMs rather than modifying existing datasets. We introduce a self-inspiration question generation mechanism to construct questions across different categories. 
Firstly, We meticulously craft one seed question and a detailed definition for each category, forming the initial question seed corpus. 
Subsequently, we incorporate these initial seed questions as examples into a meticulously designed prompt to guide GPT4-based agents in generating questions that adhere to specific category constraints. 
We limit the output to five questions in the initial rounds. These questions are manually annotated to identify new patterns and added to the seed corpus.
We continuously select representative questions into the question seed corpus to promote benchmark qualities, eventually maintained at 50 questions, serving as the test set questions for \benchmark{}.
Upon reaching 50 questions per category, we conduct manual annotations on a sample basis (30\%), with the remaining questions validated by another GPT-4 agent through a question verification process, eventually serving as the questions for \trainset{}.

\subsection{Answer Annotation}
\label{secsec:answer_annotation}
We design a self-consistency mechanism for annotating answers based on a given table and question. During the answer generation phase, we utilize three LLM agents, each employing a distinct reasoning method (TCoT, SCoT, and PoT) to generate responses. 
We introduce a voting mechanism to assess the answers generated by the different agents. We preliminarily reserve the results if the voting system identifies a valid consistency among all agents. These preliminary results are then subjected to manual review and modification to produce the final answer and its associated reasoning details. Additionally, to minimize bias in answers generated by LLMs, we enforce a strict format for all answers, retaining only the essential and accurate content, thereby avoiding any preference for model-specific answer styles.
For answers excluded due to inconsistencies, particularly those stemming from questions deemed too complex for LLMs to generate an adequate response, we randomly select 30\% of the filtered data for manual annotation and subsequently incorporate them into the dataset.
Notably, We manually annotate all answers in the \benchmark{} with no omissions and carefully scrutinize each.

\subsection{Dataset Statistic}
\label{secsec:data_satatisticcs}

\paragraph{Topics}
\benchmark{} primarily consists of numerical tables, with the largest portions derived from financial reports and data from competitive events, as illustrated in Figure \ref{fig:table_topic_distribution}, 

\paragraph{Question Categories} 
Drawing from real-world scenarios and user demands for tabular data, we devise four primary question categories: fact-checking, numerical reasoning, data analysis, and visualization, encompassing 18 subcategories, thoroughly illustrating the various challenges encountered in \simtask{} scenarios, as shown in Table \ref{table:dataset_statistic}. 
Compared to existing datasets, \benchmark{} covers a wider range of question types, as shown in Table \ref{table:comparison_with_other_dataset}, with a particular focus on data analysis and chart generation capabilities, which are notably lacking in previous datasets.

\paragraph{Reasoning Steps}
We define the complexity of the dataset by calculating the number of reasoning steps required to solve the problem. 
Figure \ref{fig:reasoning_steps} illustrates that the overall complexity of the benchmark is significantly greater than that of existing datasets, especially concerning questions related to data analysis and visualization.

\section{\ourmethod{}}
\label{sec:medthod}

\subsection{Problem Definition}
Table question answering (Table QA) can be formulated as follows: Given a semi-structured table $\mathcal{T}$, comprised of  $\mathcal{R}$ rows and $\mathcal{C}$ columns, the objective is to generate an answer $\mathcal{A}$ to a question $\mathcal{Q}$ utilizing the information contained within $\mathcal{T}$, where $\mathcal{A}$ is a set of values or entities denoted as $\{a_1, a_2, \ldots, a_k\}$, where $k \in \mathbb{N}^+$.

\subsection{Reasoning Methods}
\label{sec:reasoning_methods}
In-context learning (ICL)~\citep{dong2022survey} refers to strategies that optimize input for LLMs ($\mathcal{M}$) to generate practical outputs with a task-specific instruction ($\mathcal{I}$) and a few output examples ($\mathcal{E}$). We introduce distinct reasoning methods to fully assess the reasoning capabilities of LLMs

\paragraph{Textual Chain-of-Thought (TCoT)}
TCoT~\citep{cot} refers to a reasoning process in which LLMs incrementally derive a series of intermediate steps or sub-goals through textual prompts before generating the final answer. These intermediate steps constitute a "thought chain" that ultimately leads the model to the correct outcome. Formally, the method is:
\begin{MiddleEquation}
\begin{align}
\mathcal{M}(\mathcal{T},\mathcal{Q},\mathcal{E}) \rightarrow \{r_1, r_2, \ldots, r_k, \mathcal{A}\}
\end{align}
\end{MiddleEquation}where $r_k$ represents the $k$-th reasoning step. 

\paragraph{Symbolic Chain-of-Thought (SCoT)}
SCoT implements a methodology that utilizes Python-based instruction to facilitate logical reasoning, comprising three primary steps repeated until a definitive conclusion is derived:
\texttt{STEP-1}: Analyzing the available information to determine the next move. \texttt{STEP-2}: Generating instructions using Python programming language commands. \texttt{STEP-3}: Simulating the outcomes by executing the instructions and analyzing the results.
The entire steps can be formally framed as follows:
\begin{MiddleEquation}
\begin{align}
\mathcal{M}(\mathcal{T},\mathcal{Q},\mathcal{E}) \rightarrow \{(r_{a_1}, r_{p_1},r_{s_1}), \ldots, (r_{a_k}, r_{p_k},r_{s_k}), \mathcal{A}\}
\end{align}
\end{MiddleEquation}where $r_{a_k}$ is the analyzing step, $r_{p_k}$ is the program commands generating step, and $r_{s_k}$ is the result simulation step. 

\paragraph{Program-of-Thoughts (PoT)}
PoT~\citep{chenprogram} offers a novel approach to numerical reasoning tasks by distinctly delineating computation from reasoning. PoT decomposes the problem into programming commands $\mathcal{P}$ and utilizes a language interpreter, like Python, to compile and execute the resultant code. In contrast to SCoT, PoT enhances reasoning capabilities by actually executing generated code ($\mathcal{P}$) within a programming environment to output results, thereby implementing reasoning through structured code steps. The method can be formulated as:
\begin{MiddleEquation}
\begin{align}
\mathcal{M}(\mathcal{T},\mathcal{Q},\mathcal{E}) \rightarrow \mathcal{P} \rightarrow \mathcal{A} 
\end{align}
\end{MiddleEquation}


\subsection{Supervised Fine-Tuning}
We train \ourmethod{} by fintuning all parameters of baseline LLMs to learn from the \trainset{}. 
The training objective $\mathcal{L}_{all}$ can be described as:
\begin{MiddleEquation}
\begin{align}
    \mathcal{L}_{all} = -\sum_{n=1}^{N} \mathbb{E}_{q^{R_{n}}, a^{R_{n}} \sim \{ D^{R_{n}} \}_{n =1}^{N} } \left[ \log P(a^{R_{n}}|q^{R_{n}}; \mathcal{M}) \right]
    \label{equ:multilingual_loss}
\end{align}
\end{MiddleEquation}where $q^{R_{n}}$ and $a^{R_{n}}$ are the table-related question and answer from the dataset $D^{R_{n}}$ of reasoning method $R_{n}$, respectively. $N$ is the number of reasoning methods.

\section{Experiments}

\subsection{Implementation Details}
We meticulously design uniform style prompt templates to implement distinct reasoning methods to ensure the fairness of the evaluation. Furthermore, we impose formatting constraints on the outputs of LLMs and parse the final answers from the outputs to prevent any extraneous information from affecting the evaluation results. 
For open-source models, we operate within the \textit{transformer} environment on multiple A100 GPUs. 
For proprietary models, we employ official APIs to interact with exclusive LLMs. 
We conduct supervised finetuning of various open-source LLMs on the designated training set (\trainset{}). We utilize a cosine annealing scheduler, setting the initial learning rate at $2e^{-5}$, and conduct training over three epochs. Optimization is performed using the Adam optimizer, with a batch size of 512 and a maximum sequence length of 4096.

\subsection{LLMs}
We evaluate 34 models with sizes ranging from 7B to 110B parameters, including general/code LLMs, open-source/proprietary models, and SFT~\citep{ouyang2022training} models.
For open-source LLMs, we evaluate on Llama2s~\citep{touvron2023llama}, Llama3s~\citep{dubey2024llama}, Llama3.1s, CodeLlamas~\citep{roziere2023code}, CodeQwen1.5-7B-Chat, Qwen1.5s~\citep{bai2023qwen}, Qwen2s~\citep{qwen2}, Mistral-7B-Instruct-v0.2~\citep{jiang2023mistral}, Deepseek-Coders~\citep{guo2024deepseekcoder}, StructLMs~\citep{zhuang2024structlm}, MAP-Neo-7B-Instruct~\citep{zhang2024map}, WizardLM-13B-V1.2~\citep{xu2023wizardlm}.
For proprietary LLMs, we perform evaluation on GPTs~\citep{gpt3,gpt4} (GPT-3.5-Turbo, GPT4-Turbo, GPT4-o), Qwen-Max~\citep{qwen2}, GLM-4~\citep{chatglm}, Yi-Large~\citep{ai2024yi} and Deepseek models~\citep{bi2024deepseek} (Chat-V2, Coder-V2).
Furthermore, we finetune \ourmethod{} based on CodeQwen-7B, DeepSeekCoder-7B, Llama3-8B, Llama3.1-8B, and Qwen2-7B to further explore the Table QA capabilities of LLMs.

\subsection{Automatic Evaluation Metrics}
we adopt Rouge-L~\citep{lin-2004-rouge} to assess the quality of the generated answers by measuring the n-gram overlap with reference answers. 
In the PoT method, we enforce a specific format for the executable code outputs and evaluate the final answer with the ROUGE-L metric, ensuring alignment with other reasoning methodologies. 
Specifically, in the task of chart generation, we parse and execute code derived from LLM responses and establish rigorous test cases to assess the accuracy of the generated charts, with a particular focus on the precision of y-axis fields, employing the pass@$1$ metric \citep{chen2021evaluating} for evaluation.

\begin{table*}[htbp]
\centering
\scriptsize \selectfont
\begin{tabular}{l|cccccccccccc|ccc}
\toprule
\toprule
 & \multicolumn{3}{c}{\textbf{Fact Checking}} & \multicolumn{3}{c}{\textbf{Num-Reasoning}} & \multicolumn{3}{c}{\textbf{Data Analysis}} & \multicolumn{3}{c}{\textbf{Visualization}} &\multicolumn{3}{c}{\textbf{Overall}} \\
 \multirow{-1}{*}{} & TCoT & SCoT & PoT &  TCoT & SCoT & PoT & TCoT & SCoT & PoT & TCoT & SCoT & PoT & TCoT & SCoT & PoT \\ 
 \midrule
\textbf{Human Performance} & \multicolumn{3}{c}{\textbf{94.3}}  & \multicolumn{3}{c}{\textbf{87.1}} & \multicolumn{3}{c}{\textbf{82.1}} & \multicolumn{3}{c}{\textbf{86.3}} & \multicolumn{3}{c}{\textbf{85.91}} \\
\midrule
\multicolumn{16}{c}{\textbf{\textit{Open-source In Context Learning Methods}}} \\ 
\midrule
Llama2-7B & 34.99 & 27.47 & 3.61 & 6.70 & 4.63 & 3.95 & 14.31 & 12.49 & 1.56 & 0.00 & 0.00 & 0.00 & 12.36 & 9.95 & 2.76 \\
CodeLlama-7B & 33.06 & 12.34 & 19.44 & 5.43 & 2.99 & 13.31 & 16.16 & 17.06 & 1.79 & 0.00 & 0.00 & 0.00 & 12.30 & 9.28 & 8.85 \\
Gemma-7B & 27.63 & 10.07 & 21.62 & 6.78 & 2.91 & 10.45 & 20.33 & 11.76 & 6.74 & 0.00 & 0.00 & 2.00 & 13.96 & 6.97 & 9.81 \\
Mistral-7B & 50.45 & 40.56 & 6.25 & 8.73 & 5.77 & 2.60 & 21.99 & 21.12 & 1.19 & 0.00 & 0.00 & 0.00 & 17.86 & 15.11 & 2.35 \\
Deepseek-Coder-7B & 22.92 & 27.48 & 48.98 & 6.45 & 5.61 & 34.66 & 18.73 & 20.72 & 18.17 & 8.00 & 18.00 & 18.00 & 13.10 & 14.58 & 28.89 \\
CodeQwen1.5-7B & 30.56 & 32.94 & 0.00 & 6.24 & 5.68 & 0.00 & 27.04 & 22.47 & 0.00 & 2.00 & 0.00 & 0.00 & 16.80 & 14.85 & 0.00 \\
Qwen1.5-7B & 56.08 & 53.53 & 39.23 & 11.30 & 10.99 & 20.40 & 24.77 & 22.96 & 7.66 & 0.00 & 0.00 & 0.00 & 20.70 & 19.65 & 16.29 \\
Qwen2-7B & 57.70 & 57.52 & 0.00 & 16.09 & 16.65 & 0.76 & 24.02 & 21.50 & 0.38 & 0.00 & 4.00 & 2.00 & 22.77 & 22.26 & 0.60 \\
StructLM-7B & 47.72 & 64.06 & 13.54 & 9.55 & 19.97 & 11.48 & 19.59 & 23.83 & 4.38 & 0.00 & 0.00 & 0.00 & 17.06 & 25.21 & 8.30 \\
MAP-Neo-7B & 32.70 & 33.22 & 0.00 & 7.23 & 6.46 & 0.00 & 21.85 & 14.38 & 0.44 & 0.00 & 0.00 & 4.00 & 15.26 & 12.03 & 0.40 \\
Llama3-8B & 38.32 & 72.53 & 13.94 & 22.02 & 17.33 & 19.50 & 30.15 & 30.75 & 9.31 & 0.00 & 0.00 & 10.00 & 25.71 & 27.59 & 14.43 \\
Llama3.1-8B & 47.89 & 36.29 & 30.38 & 11.26 & 13.77 & 17.24 & 15.78 & 14.82 & 8.86 & 8.00 & 0.00 & 8.00 & 16.76 & 15.81 & 14.88 \\
Llama2-13B & 48.47 & 32.69 & 3.03 & 15.83 & 6.79 & 4.48 & 22.04 & 17.16 & 3.19 & 0.00 & 0.00 & 0.00 & 20.86 & 13.25 & 3.61 \\
StructLM-13B & 26.28 & 64.49 & 1.04 & 12.30 & 17.38 & 0.00 & 20.70 & 18.41 & 0.28 & 0.00 & 0.00 & 0.00 & 16.35 & 21.94 & 0.21 \\
WizardLM-13B & 53.93 & 46.01 & 8.33 & 13.79 & 16.52 & 14.79 & 22.61 & 20.16 & 3.73 & 0.00 & 0.00 & 4.00 & 20.75 & 20.23 & 9.12 \\
Qwen1.5-14B & 40.83 & 61.92 & 44.38 & 10.29 & 15.01 & 28.20 & 22.99 & 29.24 & 10.33 & 2.00 & 8.00 & 2.00 & 18.03 & 25.14 & 21.48 \\
Qwen1.5-32B & 64.99 & 67.86 & 49.01 & 19.13 & 21.15 & 34.01 & 24.27 & 28.29 & 17.43 & 4.00 & 8.00 & 8.00 & 25.38 & 28.30 & 27.79 \\
Deepseek-Coder-33B & 48.27 & 54.34 & 33.12 & 9.41 & 12.69 & 32.60 & 9.09 & 21.70 & 19.97 & 0.00 & 0.00 & 24.00 & 13.01 & 19.92 & 27.20 \\
CodeLlama-34B & 64.39 & 58.28 & 5.90 & 13.10 & 13.30 & 4.20 & 19.23 & 15.28 & 0.53 & 0.00 & 0.00 & 2.00 & 20.24 & 18.19 & 2.88 \\
StructLM-34B & 19.10 & 30.21 & 27.74 & 15.36 & 9.03 & 14.45 & 20.74 & 17.92 & 5.38 & 0.00 & 0.00 & 2.00 & 16.93 & 14.37 & 11.61 \\
Mixtral-8x7B & 54.54 & 56.01 & 35.86 & 16.80 & 16.05 & 26.23 & 24.69 & 25.67 & 13.96 & 2.00 & 0.00 & 6.00 & 23.14 & 23.24 & 21.32 \\
Qwen1.5-72B & 71.27 & 67.03 & 33.16 & 19.01 & 16.68 & 20.85 & 26.63 & 27.33 & 13.03 & 2.00 & 8.00 & 14.00 & 26.66 & 25.80 & 18.65 \\
Qwen2-72B & 72.50 & 71.13 & 56.37 & 36.97 & 31.81 & 41.33 & 32.20 & 31.85 & 22.36 & 20.00 & 14.00 & 12.00 & 38.13 & 35.14 & 33.91 \\
Qwen1.5-110B & 74.87 & 69.80 & 53.55 & 29.81 & 23.33 & 36.83 & 27.34 & 29.32 & 18.38 & 14.29 & 12.00 & 24.00 & 32.81 & 30.10 & 30.77 \\
Llama3-70B & 73.88 & 75.44 & 60.64 & 37.64 & 28.87 & 36.59 & 37.47 & 34.06 & 26.11 & 4.00 & 6.00 & 10.00 & 39.59 & 34.48 & 33.59 \\
Llama3.1-70B & 76.32 & 77.65 & 59.05 & 44.89 & 38.93 & 34.04 & 35.88 & 33.87 & 23.15 & 26.00 & 6.00 & 34.00 & 43.85 & 39.22 & 32.52 \\
\midrule
\multicolumn{16}{c}{\textbf{\textit{Close-source In Context Learning Methods}}} \\ 
\midrule
GPT-3.5-Turbo & 59.95 & 75.68 & 60.92 & 23.45 & 23.16 & 42.09 & 34.40 & 32.54 & 30.25 & 10.00 & 4.00 & 38.00 & 30.85 & 31.41 & 39.34 \\
Qwen-Max & 70.48 & 68.21 & 50.42 & 32.83 & 25.06 & 32.80 & 27.87 & 30.98 & 19.41 & 18.00 & 8.00 & 30.00 & 34.26 & 31.04 & 29.39 \\
Yi-Large & 71.41 & 66.08 & 13.19 & 40.18 & 23.20 & 15.25 & 29.22 & 22.59 & 5.97 & 26.00 & 26.00 & 6.00 & 38.57 & 27.82 & 10.90 \\
GLM-4 & 67.93 & 73.59 & 31.49 & 34.01 & 26.18 & 25.46 & 30.47 & 31.54 & 25.34 & 8.00 & 14.00 & 22.00 & 34.80 & 32.76 & 25.92 \\
Deepseek-Chat-V2 & 72.41 & 69.89 & 57.48 & 50.07 & 38.96 & 45.96 & 38.07 & 34.44 & 30.37 & 40.00 & 24.00 & 46.00 & 47.22 & 39.63 & 41.07 \\
Deepseek-Coder-V2 & 73.90 & 70.17 & 63.00 & 47.22 & 38.24 & 47.26 & 33.09 & 31.82 & 31.56 & 40.00 & 26.00 & 44.00 & 44.26 & 38.57 & 42.75 \\
GPT-4-Turbo & 75.92 & 77.62 & 70.08 & 53.01 & 44.31 & 49.31 & 41.03 & 36.52 & 34.63 & 62.00 & 32.00 & 48.00 & 51.32 & 44.26 & 45.69 \\
GPT-4o & 72.63 & 71.01 & 62.31 & 54.46 & 42.20 & 47.83 & 38.90 & 34.65 & 30.03 & 56.00 & 38.00 & 54.00 & 50.39 & 42.22 & 42.92 \\
\midrule
\multicolumn{16}{c}{\textbf{\textit{Open-Source Fine-Tuning Methods}}} \\ 
\midrule
TableLLM$_{\text{CodeQwen-7B}}$ & 62.90 & 66.94 & 4.86 & 24.86 & 14.90 & 12.04 & 31.49 & 30.52 & 16.08 & 36.00 & 26.00 & 36.00 & 32.16 & 27.13 & 14.21 \\
TableLLM$_{\text{Dpsk-Coder-7B}}$ & 69.23 & 63.15 & 7.12 & 35.87 & 21.20 & 14.61 & 31.75 & 29.62 & 21.04 & 36.00 & 18.00 & 30.00 & 37.76 & 28.80 & 17.19 \\
TableLLM$_{\text{Llama3.1-8B}}$  & 68.15 & 65.17 & 25.67 & 30.51 & 17.86 & 28.64 & 33.47 & 30.20 & 19.77 & 24.00 & 18.00 & 44.00 & 35.29 & 27.70 & 25.76 \\
TableLLM$_{\text{Llama3-8B}}$  & 62.13 & 64.46 & 15.07 & 29.42 & 16.73 & 12.68 & 30.21 & 29.45 & 17.53 & 26.00 & 20.00 & 28.00 & 32.89 & 26.97 & 15.83 \\
TableLLM$_{\text{Qwen2-7B}}$  & 71.05 & 62.34 & 10.59 & 37.25 & 19.95 & 10.34 & 32.60 & 31.95 & 18.76 & 24.00 & 20.00 & 26.00 & 38.32 & 29.09 & 14.54 \\
\midrule
\bottomrule
\end{tabular}
\caption{
The main results of advanced LLMs on TableBench are presented alongside human performance. All methods involving code generation and computation, particularly in the chart generation task, execute code only once to derive the final answer. The overall results represent a weighted average of performance across different categories.}
\label{tab:main_results}
\end{table*}

\subsection{Main Results} 
Table \ref{tab:main_results} showcases the main results of over 30 advanced advanced LLMs on the \benchmark{}.
GPT-4 outperforms other models in numerous tasks, demonstrating superior performance across complex reasoning scenarios. Particularly in numerical computation and analytical tasks, GPT-4 maintains a commendable level of performance.
\ourmethod{} finetuned on the open-source models with \trainset{} achieves a performance level comparable to GPT-3.5, significantly validating the effectiveness of our training data.
Despite these advancements, humans still surpass all LLMs in these tasks. Nevertheless, certain advanced LLMs, especially those employing proprietary approaches, demonstrate potential in these scenarios.
However, complex reasoning environments on tabular data still remain challenges.

\begin{table}[bp]
\centering
\scriptsize \selectfont
\begin{tabular}{l|ccc}
\toprule
 &\textbf{Auto Metric} & \textbf{GPT-4 Eval} & \textbf{Human Eval}  \\
\midrule
GPT-3.5-Turbo & 30.87 & 32.84 & 34.12  \\
Qwen-Max & 34.29 & 36.12 & 37.12  \\
Yi-Large & 38.56 & 43.12 & 41.23  \\  
GLM-4 & 34.82 & 38.60 & 39.21 \\
Deepseek-Chat-V2 & 47.24 & 48.31 & 50.12 \\
Deepseek-Coder-V2 & 44.92 & 44.92 & 46.13 \\ 
GPT-4-Turbo & 51.30 & 52.82 & 54.02 \\
GPT-4o & 50.53 & 54.18 & 53.19 \\
\midrule
\textbf{\makecell[c]{PCC with Auto Metric}} & 1.000 & 0.981 & 0.995 \\
\bottomrule
\end{tabular}
\caption{We performed a consistency test of evaluation methods for advanced LLMs on TCoT performance}
\label{tab:human_eval}
\end{table}

\paragraph{Category Analysis}
Experimental results in Table \ref{tab:main_results} reveal that most models perform commendably in fact-based reasoning tasks, indicating their proficiency in this area. However, challenges arise in numerical reasoning tasks due to the complexity of mathematical computations, especially complex calculations such as aggregation, which require multiple intermediate steps to reach the final answer. Data analysis tasks necessitate more intricate and comprehensive analytical skills, such as using correlation coefficients to analyze model relationships and employing linear regression functions to predict future trends, thereby imposing higher demands on the overall reasoning abilities of LLMs. The task of chart generation poses the greatest challenge, requiring significant coding skills and strict adherence to instructions. 
Notably, smaller-sized models exhibit significant deficiencies in chart generation tasks, highlighting their limitations in utilizing code to handle complex tasks.

\paragraph{Reasoning Methods Analysis}
As illustrated in Table \ref{tab:main_results}, those methods incorporating reasoning steps demonstrate a clear advantage on TableBench compared to methods that derive conclusions directly.
The TCoT method exhibits stable and superior performance across various dimensions.
The PoT method delivers commendable results in purely numerical computations, particularly in chart generation, but falls short in textual reasoning.
We investigate the factors contributing to the suboptimal performance of the PoT method and find that the code execution success rate constrains the performance, as we only conduct a single generation and execute the code without employing any strategy for code correction. 
Even for the best-performing GPT4-Turbo, the executable code ratio is only 78.67\%. This indicates that the PoT method requires LLMs with significant code-generation capabilities and instruction-following ability. However, it also underscores the substantial potential of the PoT method.
Conversely, the SCoT method adapts effectively in scenarios requiring a combination of numerical and textual reasoning, such as analytical tasks, achieving a balanced yet modest overall performance. The performance of SCoT falls short of expectations due to its reliance on simulated outcomes rather than executing actual code.

\subsection{Consistency of Evaluation Methods}
Despite constraints imposed on the output format and the standardization of ground truth annotations, the ROUGE-L metric may not fully capture the real performance due to the inherent flexibility in the outputs of LLMs. 
Both GPT-4 and human judgment are conducted, as shown in Table \ref{tab:human_eval}, to assess this potential bias. The Pearson Correlation Coefficient~\citep{cohen2009pearson} (PCC) is adopted to analyze the consistency across different evaluation methods. The results, as presented in the table, indicate a high level of agreement among these evaluating methods, demonstrating that the constraints are effective and our metric accurately reflects the real performance of LLMs on the \benchmark{}.


\section{Further Analysis}

\begin{figure}[tp]
\centering
\includegraphics[]{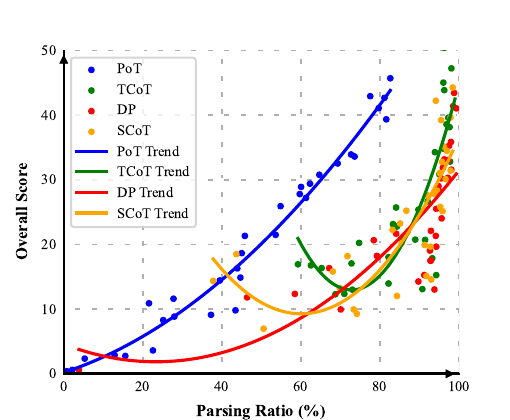}    
\caption{The impact of the parsing ratio on the overall score, where the parsing ratio is defined as the proportion of responses generated by the LLM that can be successfully parsed according to predetermined instructions.}
\label{fig:parse_efficiency}
\end{figure}

\subsection{Instruct Following Analysis}
We observe that the performance trends of small-size LLMs across different reasoning methods differ from those observed in large-size models in Table \ref{tab:main_results}. 
In further analysis, we introduce a comparison with non-reasoning methods, specifically focusing on Direct Prompting (DP), which provides solutions directly without intermediate reasoning steps. 
We find that the non-reasoning method (DP) performs better on small-size LLMs than reasoning-based methods. 
As shown in Figure \ref{fig:parse_efficiency}, most models exhibit good instruction-following capabilities with the DP method due to the simpler instructions to follow. Conversely, small-size LLMs perform significantly worse with the PoT method, mainly due to their insufficient code generation capabilities, resulting in a lower rate of executable code generation. Additionally, the iterative symbolic reasoning steps required by the SCoT method pose considerable challenges for small-scale models.

In comparison to the DP, SCoT, and TCoT methods in Figure \ref{fig:parse_efficiency}, the data points on the left side of the quadratic curve show that at low parsing ratios, the overall score increases as the parsing ratio decreases, suggesting that certain models (e.g., StructLLM), possess strong table understanding capabilities but exhibit weaker instruction-following abilities. This may be attributed to differences in the instruction format during instruction tuning compared to the format we employ. 
The right side of the quadratic curve reveals that despite the strong instruction-following performance of the DP method, the non-reasoning DP method faces a clear performance ceiling. In contrast, reasoning-based methods show significant potential for improvement.
The curve of the PoT highlights the substantial potential of the PoT to enhance the overall score by increasing the parsing rate.

\begin{figure}[tp]
\centering
\includegraphics[]{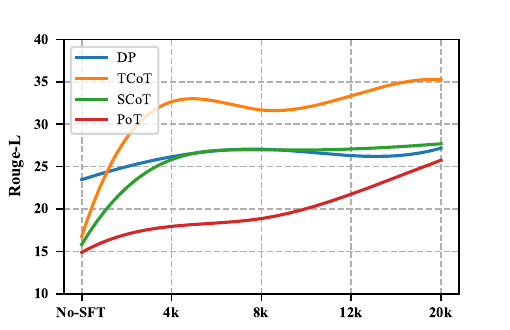}
\caption{TableInstruct data efficient on TableLLM$_{\text{Llama3.1-8B}}$}
\label{fig:instruct_size_exp}
\end{figure}

\subsection{Data Efficiency of \trainset{}}
In this section, we discuss the data efficiency of \trainset{} on the SFT process. We construct datasets of varying sizes by sampling from \trainset{} with sampling rates ranging from 0.2 to 0.6. Figure \ref{fig:instruct_size_exp} visually depicts the relative performance at different sampling rates. Surprisingly, with only 60\% of the samples, the model retains over 90\% of the performance of the complete dataset. 
We observe that the Llama-3-8B model requires fewer than 4,000 samples to surpass the performance of Qwen1.5-70B on the dataset, demonstrating that the \trainset{} corpus significantly enhances tabular reasoning in smaller models.
The full data provides the highest knowledge coverage, enabling the model to achieve optimal overall performance, comparable to GPT-3.5, with inference costs being only a fraction, indicating the high efficiency of \trainset{}.


\section{Related Work}
\label{sec:relatedwork}

Table QA~\citep{mueller2019answering,jin2022survey} has grown substantially, driven by the development of robust datasets that engage advanced algorithms in the tasks of semantic comprehension~\cite{DBLP:conf/coling/HuangMLHZ0Z24, DBLP:journals/corr/abs-2311-11268, DBLP:journals/corr/abs-2402-11100,bai2023qwen,qwen2,li1,li2,li3}. These datasets function as significant milestones for enhancing table-centric semantic understanding.
WTQ~\citep{pasupat2015compositional}, SQA~\citep{iyyer2017search}, and TabFact~\citep{chen2019tabfact} set the cornerstone for Table QA research. They furnish benchmarks founded on question-answer pairs predicated on HTML tables sourced from Wikipedia. 
However, these datasets rely heavily on specific cell content from the table to formulate answers, which can not fully represent the multi-dimensional queries posed in real-world scenarios.

Acknowledging this incongruity, some datasets have been introduced to bridge the gap. ToTTo~\citep{parikh2020totto}, OTTQA~\citep{chen2020open}, and FeTaQA~\citep{nan2022fetaqa} step into the fore by providing free-form QA datasets. These datasets challenge models to generate answers that go beyond the table's explicit content, thereby enhancing model performance to align with the free-form nature of real-world questions.
FinQA~\citep{chen2021finqa} and AIT-QA~\citep{katsis2022ait} lay emphasis on numeric-focused queries. These datasets predominantly target financial tables, suggesting complex reasoning challenges that necessitate models to not only interpret but also to compute and extract nuanced information precisely.
Further diversifying the landscape, datasets such as WikiSQL~\citep{zhong2017seq2sql}, Spider~\citep{yu2018spider}, and Bird~\citep{li2023can} introduce logical expressions as supervisory signals to train Table QA models, discreting reasoning capabilities through logic-based problem-solving.
Despite the significant advancements made by LLMs in TableQA~\citep{DBLP:conf/acl/LiZLLLSWLCZ22, singha2023tabular, li2023table, lei2023tableqakit,he2023text2analysisbenchmarktablequestion}, there is still a critical need for benchmarks that reflect the reasoning complexity encountered in real-world tabular data scenarios. \benchmark{}, a comprehensive and complex benchmark, incorporates real-world complexities into its evaluation scenarios, effectively addressing the limitations of existing benchmarks

\section{Conclusion}
In this work, we introduce \benchmark{}, a comprehensive and complex benchmark designed to evaluate a broad spectrum of tabular skills.
It encompasses 886 question-answer pairs across 18 distinct capabilities, significantly contributing to bridging the gap between academic benchmarks and real-world applications. 
We evaluate 30+ models with various reasoning methods on \benchmark{} and provide a training set \trainset{} that enables \ourmethod{} to achieve performance comparable to ChatGPT. 
Despite these advancements, even the most advanced model, GPT-4, still lags significantly behind human performance on \benchmark{}, underscoring the challenges of tabular tasks in real-world applications.

\section*{Limitations}
We acknowledge the following limitations of this study: 
(1) This paper mainly focuses on the reasoning complexity of table questions, which does not extensively explore the inherent complexities of the tables themselves.
(2) Tabular data in image formats, which are also prevalent in real-world applications, are not discussed in this paper. 

\section{Acknowledgments}
This work was supported in part by the National Natural Science Foundation of China (Grant Nos. 62276017, 62406033, U1636211, 61672081), and the State Key Laboratory of Complex \& Critical Software Environment (Grant No. SKLCCSE-2024ZX-18).

\bibliography{custom-fix-sim}

\end{document}